\def\year{2019}\relax
\documentclass[letterpaper]{article} %DO NOT CHANGE THIS
\usepackage{aaai19}  %Required
\usepackage{times}  %Required
\usepackage{helvet}  %Required
\usepackage{courier}  %Required
\usepackage{url}  %Required
\usepackage{graphicx}  %Required
\usepackage{amsmath}
\usepackage[utf8]{inputenc}
\usepackage{color}
\usepackage{cases}
\usepackage{nicefrac}
\usepackage{tabu}
\usepackage{booktabs}
\usepackage{multirow}
\usepackage{subcaption}
\usepackage{caption}
\usepackage{soul}

\usepackage[dvipsnames]{xcolor}
\usepackage{amsmath,amssymb}%\usepackage{mathbbol}
\usepackage{array}
\usepackage[english]{babel}
\usepackage{algorithm}
\usepackage{algpseudocode}
\usepackage{hyperref}

\begin{document}

\title{Backbone Can Not be Trained at Once: \\
Rolling Back to Pre-trained Network for Person Re-identification}

\author{
  Youngmin Ro\textsuperscript{1}
  \,
  Jongwon Choi\textsuperscript{2}
  \,
  Dae Ung Jo\textsuperscript{1}
  \,
  Byeongho Heo\textsuperscript{1}
   \,
  Jongin Lim\textsuperscript{1}
   \,
  Jin Young Choi\textsuperscript{1}
  \\
  \small{\texttt{\{treeoflife, mardaewoon, bhheo, ljin0429, jychoi\}@snu.ac.kr},\; 
  \texttt{jw17.choi@samsung.com}}\\
  \textsuperscript{1}Department of ECE, ASRI, Seoul National University, Korea\\
  \textsuperscript{2}Samsung SDS, Korea\\
}

\nocopyright
\maketitle
\begin{abstract}
In person re-identification (ReID) task, because of its shortage of trainable dataset, it is common to utilize fine-tuning method using a classification network pre-trained on a large dataset. However, it is relatively difficult to sufficiently fine-tune the low-level layers of the network due to the gradient vanishing problem. In this work, we propose a novel fine-tuning strategy that allows low-level layers to be sufficiently trained by rolling back the weights of high-level layers to their initial pre-trained weights. Our strategy alleviates the problem of gradient vanishing in low-level layers and robustly trains the low-level layers to fit the ReID dataset, thereby increasing the performance of ReID tasks. 
The improved performance of the proposed strategy is validated via several experiments. Furthermore, without any add-ons such as pose estimation or segmentation, our strategy exhibits state-of-the-art performance using only vanilla deep convolutional neural network architecture.
Code is available at \href{https://github.com/youngminPIL/rollback}{\url{https://github.com/youngminPIL/rollback}}

\end{abstract}

\section{Introduction}
Person re-identification (ReID) refers to the tasks connecting the same person, for instance, a pedestrian, among multiple people detected in non-overlapping camera views.
Different camera views capture pedestrians in various poses  with different backgrounds, which interferes with the ability to correctly estimate the similarity among pedestrian candidates. 
These obstacles makes it difficult to recognize the identities of numerous pedestrians robustly by comparing them with a limited number of person images with known identities.
Furthermore, it is infeasible to obtain large training datasets sufficient to cover the appearance variation of pedestrians, making the ReID problem difficult to be solved.
When sufficient training data is not available, it is a common approach to fine-tune the network pre-trained by another large dataset (\textit{e.g.}, ImageNet) which contains abundant information.
The fine-tuning approach results in better performance than the approaches in which networks are trained from randomly initialized parameters.
This is a practical approach used in many research areas~\cite{ref:faster,ref:FCN} to avoid the problem of overfitting.
Likewise, the previous ReID algorithms~\cite{ref:MLFN,ref:DuATM,ref:SVDnet} have utilized the fine-tuning approach. 
Most of recent works in ReID research have attempted to utilize semantic information such as pose estimation~\cite{ref:pose_alined,ref:AACN,ref:Pose_sensitive}, segmentation mask~\cite{ref:MaskreID}, and semantic parsing~\cite{ref:human_parsing} to improve the accuracy of ReID by considering the additional pedestrian contexts.

In contrast to the previous studies, we are interested in incrementally improving the performance of ReID by enhancing the basic fine-tuning strategy applied to the pre-trained network.
A few attempts have been made to improve learning methods by the ways designing a new loss function or augmenting data in a novel way~\cite{ref:mutualReID,ref:beyond_triplet,ref:pert_erasing,ref:SVDnet}. 
However, there has been no research on improving the learning method to consider the characteristics of each layer filter.

% \input{input_fig/cover}
%%%%%%%%%%%%%%%%%%%%%%%%%%%%%%%%%%%%%%%%%%%%
\begin{figure}[t]
\centering
{\includegraphics[width=9.0cm]{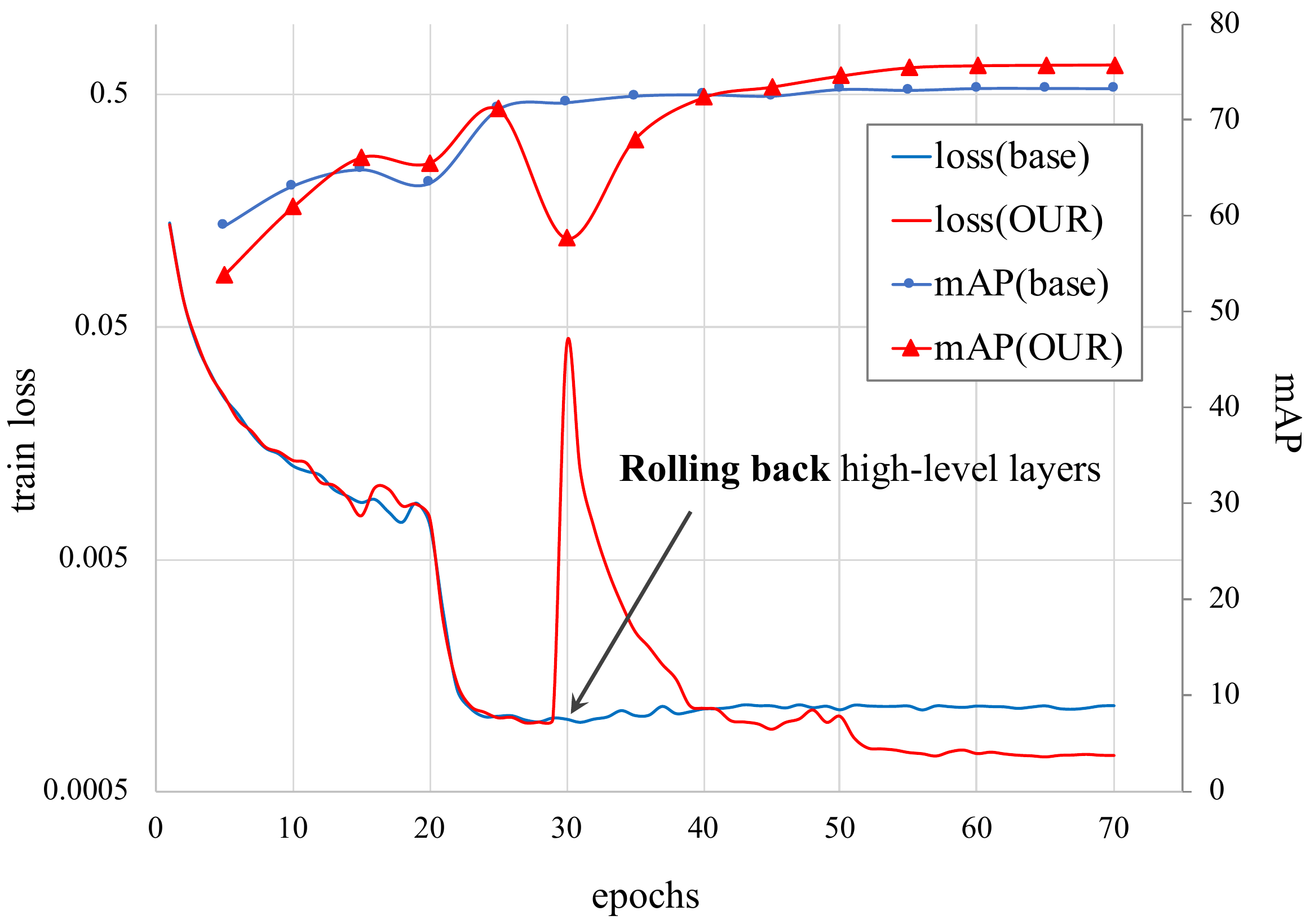}}
    \caption{Training loss and mAP graph changed by introducing our learning strategy. `base' means that the network is trained by basic strategy. In our method, the training loss escapes from local minimum and the mAP accuracy increases by utilizing the rolling-back scheme.}
    \label{fig:cover}
\end{figure}
%%%%%%%%%%%%%%%%%%%%%%%%%%%%%%%%%%%%%%%%%%%%
Before suggesting our novel fine-tuning strategy for ReID, we first empirically analyze the importance of fine-tuning low-level layers for ReID problems. According to related research~\cite{ref:zfnet,ref:understand_deepimage}, the low-level layers concentrate on details of appearance to discriminate between samples while the high-level layers contain semantic information. Thus, we need to sufficiently fine-tune the low-level layers to improve the discriminant power for the specific class ‘person’ in ReID because the low-level layers of the pre-trained network include detailed information on numerous classes. 
However, since the gradients delivered from high-level layers to low-level layers are reduced through back-propagation, the low-level layers suffer from a gradient-vanishing problem, which causes early convergence of the entire network before the low-level layers are trained sufficiently.

To solve this problem, we propose a novel fine-tuning strategy in which a part of the network is intentionally perturbed when learning slows down. The proposed fine-tuning strategy can recover the vanished gradients by rolling back the weights in the high-level layers to their pre-trained weights, which provides an opportunity for further tuning of weights in the low-level layers. 
As shown in Figure~\ref{fig:cover}, the proposed fine-tuning strategy allows the network to converge to a minimum in a basin with better generalization performance than the conventional fine-tuning method.
We validate the proposed method that uses no add-on schemes via a number of experiments, and the method outperforms state-of-the-art ReID methods appending additional context to the basic network architecture.
Furthermore, we apply the proposed learning strategy to the fine-grained classification problem, which validates its generality for various computer vision tasks.

\section{Related Work}
Traditionally, the ReID problem has been solved by using a metric learning method~\cite{ref:metric_learning_reID} to narrow the distance among the images of the same person. 
Clothing provides an important hint in the ReID task, and some approaches~\cite{ref:color_hist_ReID,ref:color_hist_wacv} have used color-based histograms.
With the development of deep learning, many ReID methods to learn discriminative features by deep architectures appear, which dramatically increases the ReID performance~\cite{ref:SVDnet,ref:harmonious,ref:defense_triplet}.
Recently, the state-of-the-art  approaches~\cite{ref:DuATM,ref:MaskreID,ref:CamAug} have also used the advanced deep architecture, especially pre-trained on ImageNet~\cite{ref:imagenet}, as a backbone network.

\subsubsection{Add-on semantic information method in ReID}
To increase the performance, many recent works based on the deep architectures have tried to consider additional semantic information such as poses of pedestrians and attention masks.
One of the most popular approaches is to use the off-the-shelf pose estimation 
algorithms~\cite{ref:openpose,ref:deepcut_pose}
to tackle the misaligned poses of the candidate pedestrians. 
In \cite{ref:pose_driven}, using the pose information, Su~\textit{et}~\textit{al}
aligned each part of a person, producing pose-normalized input to deal with the problem of the deformable variation of the ReID object.
Sarfraz~\textit{et}~\textit{al}.~\cite{ref:Pose_sensitive} proposed a view predictor network that distinguishes the front, back, and sides of a person using pose information.
In addition to using the pose estimation algorithms, there was a method~\cite{ref:MaskreID} which embeds a 4-channel input by concatenating 3-channels of RGB input image and one channel of segmentation mask. 
Likewise, an algorithm~\cite{ref:human_parsing} uses semantic parsing masks rather than whole body mask.
In~\cite{ref:PoseNormal_GAN}, they generate a realistic pose-normalized image. The synthesized image can be used as training data because the label is preserved.
\cite{ref:AACN} proposed attention-aware composition network. They pointed out the conventional methods using pose information based on rigid body regions such as rectangular RoI. They obtained non-rigid parts through connectivity information between the human joints and matched them individually. 
In contrast to the previous ReID methods, we target on improving the training method itself without any additional semantic information or extra architecture.

\subsubsection{Advanced fine-tuning methods}
There are other studies to improve learning methods on pre-trained networks.
Li and Hoiem~\cite{ref:learning_without_forgetting} suggested a method which can learn a new task without forgetting the existing tasks in transfer learning.
In \cite{ref:do_better_imagenet}, Kornblith~\textit{et}~\textit{al}. analyzed a conventional fine-tuning method, which concluded that the state-of-the-art ImageNet architecture yields state-of-the-art results over many tasks.
In the ReID task, several methods have improved learning strategy on pre-trained networks.
The quadruplet loss was proposed in \cite{ref:beyond_triplet}. 
In this research, Chen \textit{et}~\textit{al}. have developed an improved version of triplet losses, which does not only make the inter-class close but also add a negative sample, making the distance in the intra-class much longer.
In \cite{ref:mutualReID}, Zhang \textit{et}~\textit{al} were inspired by the distillation method~\cite{ref:distillation} between teacher and student networks and proposed a learning method based on co-student networks which can be trained without teacher network. 
However, there has been no research considering the fine-tuning characteristics for the ReID problem. 
In this paper, we propose a novel fine-tuning strategy adapted to the ReID task, which takes into account the layer-by-layer characteristic of the network.

\section{Methodology}

%%%%%%%%%%%%%%%%%%%%%%%%%%%%%%%%%%%%%%%%%%%%%%%%%
\begin{figure}
\centering
{\includegraphics[width=8.0cm]{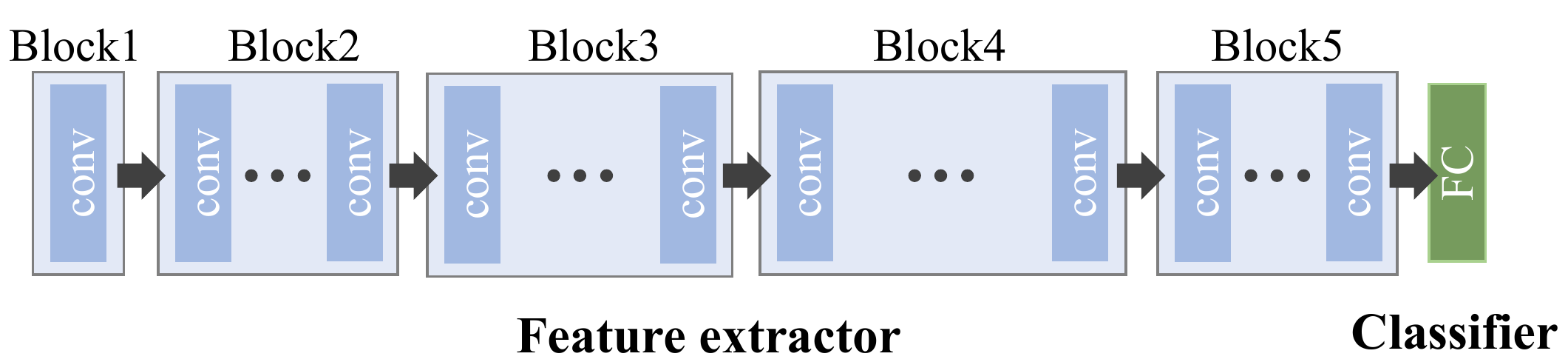}}
    \hfill    
    \caption{The description of the network: ResNet-34, ResNet-50 and ResNet-101 are utilized as a feature extractor. The classifiers are re-defined for each ReID dataset.}
    \label{fig:network}
\end{figure}
%%%%%%%%%%%%%%%%%%%%%%%%%%%%%%%%%%%%%%%%%%%%%%%%%

In this section, we first analyze the conventional fine-tuning strategy to determine which layer is insufficiently trained for ReID problems. Based on the analysis, we propose a new fine-tuning strategy that alleviates the vanishing gradient in the poorly trained layers, consequently improving the generalization performance of the fine-tuned network.

\subsection{Overall framework} 
Before describing the empirical analysis and the proposed fine-tuning strategy, we first introduce an overall framework including a network architecture with its training and testing processes.
The notations defined in this section are used in the following sections.

\subsubsection{Architecture}
In this paper, we use a classification-based network~\cite{ref:ReID_review} that determines the entire identity label as a class. 
We assume that the deep convolutional neural network consists of two components: a feature extractor and a classifier.
The feature extractor is composed of multiple convolutional layers and the classifier consists of several fully-connected (FC) layers.
As the feature extractor, we utilize convolutional layers of pre-trained ResNet~\cite{ref:resnet}, which are widely used in many ReID algorithms~\cite{ref:SVDnet,ref:CamAug,ref:PoseNormal_GAN}.
The three structures ResNet-34, ResNet-50, and ResNet-101 are used for the feature extractor to show the generality of the proposed fine-tuning strategy.
According to the resolution of the convolutional layers, the feature extractor can be partitioned into five blocks where each block contains several convolutional layers of the same resolution. 
The five blocks of ResNet-34, ResNet-50, and ResNet-101 contain $\{1, 6, 8, 12, 6\}$, {$\{1, 9, 12, 18, 9\}$}, and $\{1, 9, 12, 69, 9\}$ convolutional layers, respectively.
Following feature extraction, a feature vector is obtained by a global average pooling layer that averages the channel-wise values of the feature map resulting from the last convolutional layer. 
The resulting feature vector is a 2048-D vector for ResNet-50 and ResNet-101 and a 512-D vector for ResNet-34.
The network infers the identity of the input sample by feeding the feature vector obtained from the feature extractor into the classifier.
The classifier is newly defined in the order of 512-D FC layer, batch normalization, leaky-rectified linear unit, and FC layer with $L$-dimension, where $L$ is the number of identities in the training set and varies between datasets. Following the last FC layer, a soft-max layer is located.

\subsubsection{Training process} 
We train the network to classify the identities of training samples based on cross-entropy loss. 
The weight parameters to be trained are denoted by $\theta \equiv \{\theta_{1},...,\theta_{N},\theta_{FC}\}$, where $\theta_{n}$ and $\theta_{FC}$ are weight parameters of $n$-th block and FC layers, respectively.
Given $N$ training samples $\{{\mathbf{x}_i}\}_{i=1}^{N}$ with $L$ identities and the corresponding one-hot vectors $\{{\mathbf{y}_i}\}_{i=1}^{N}$ where $\mathbf{y}_i \in \{0,1\}^{L\times 1}$, the probability that $ \mathbf{x}_i $ corresponds to each label is calculated as:
\begin{equation}
\label{eq:forwarding}
p(\mathbf{x}_i|\theta)= \mathcal{C}(\mathcal{F}(\mathbf{x}_i|\theta_{1},.,\theta_{N})|\theta_{FC}),
\end{equation}
\noindent where $p(\mathbf{x}_i|\theta) \in \mathbb{R}^{L\times 1}$, $\mathcal{F}(\mathbf{x}_i|\theta_{1},.,\theta_{N})$ denotes feature extractor for $\mathbf{x}_i$ with $\theta_{1},.,\theta_{N}$, and $\mathcal{C}(~\cdot~|\theta_{FC})$ denotes a classifier with $\theta_{FC}$. 
The cross-entropy loss between the estimated $p(\mathbf{x}_i|\theta)$ and $\mathbf{y}_i$ is calculated as follows:

\begin{equation}
\label{eq:ce-loss}
\mathcal{L}(\mathbf{x}_i, \mathbf{y}_i, \theta) = - \frac{1}{N} \sum_{i=1}^{N}  \mathbf{y}_i^T \log{p(\mathbf{x}_i|\theta)}.
\end{equation}
In the training process, a stochastic gradient descent method is used to train $\theta$ by minimizing Eq.~\ref{eq:ce-loss}.

\subsubsection{Testing process} 
The identities given to the testing set are completely different than the identities in the training set. 
Thus, the classifier trained in the training process cannot be used for the testing process. 
To find correspondence between pedestrian candidates without using the classifier, we estimate the similarity of two pedestrians based on the distance between the feature vectors of each pedestrian extracted from the trained feature extractor.
To evaluate the performance, the testing set is divided into a query set and a gallery set with $M_q$ and $M_g$ samples, respectively. 
The samples of the query and gallery sets are denoted by $\{\mathbf{x}_{q,i}\}_{i=1}^{M_q}$\ and $\{\mathbf{x}_{g, j}\}_{j=1}^{M_g}$, respectively.
Each sample in the query set is a person of interest, which should be matched to the candidate samples in the gallery set. 

The distance between $\mathbf{x}_{q,i}$ and $\mathbf{x}_{g,j}$ is calculated by L-2 norm as follows:
\begin{align}
\boldsymbol{q}^{(i)} &= \mathcal{F}(\mathbf{x}_{q,i}|\theta_{1},...,\theta_{N}) \\ \mathbf{g}^{(j)} &= \mathcal{F}(\mathbf{x}_{g,j}|\theta_{1},...,\theta_{N}) \\
s_{i,j} &= {|| \boldsymbol{q}^{(i)} - \mathbf{g}^{(j)}||}_2^2 .
\end{align}
The identity of the gallery sample with the lowest distance $s_{i,j}$ is determined as the identity of the $i$-th query sample.
%%%%%%%%%%%%%%%%%%%%%%%%%%%%%%%%%%%%%%%%%%%%%%%%%%%%
\begin{figure}[t]
\centering
{\includegraphics[width=1.0\linewidth]{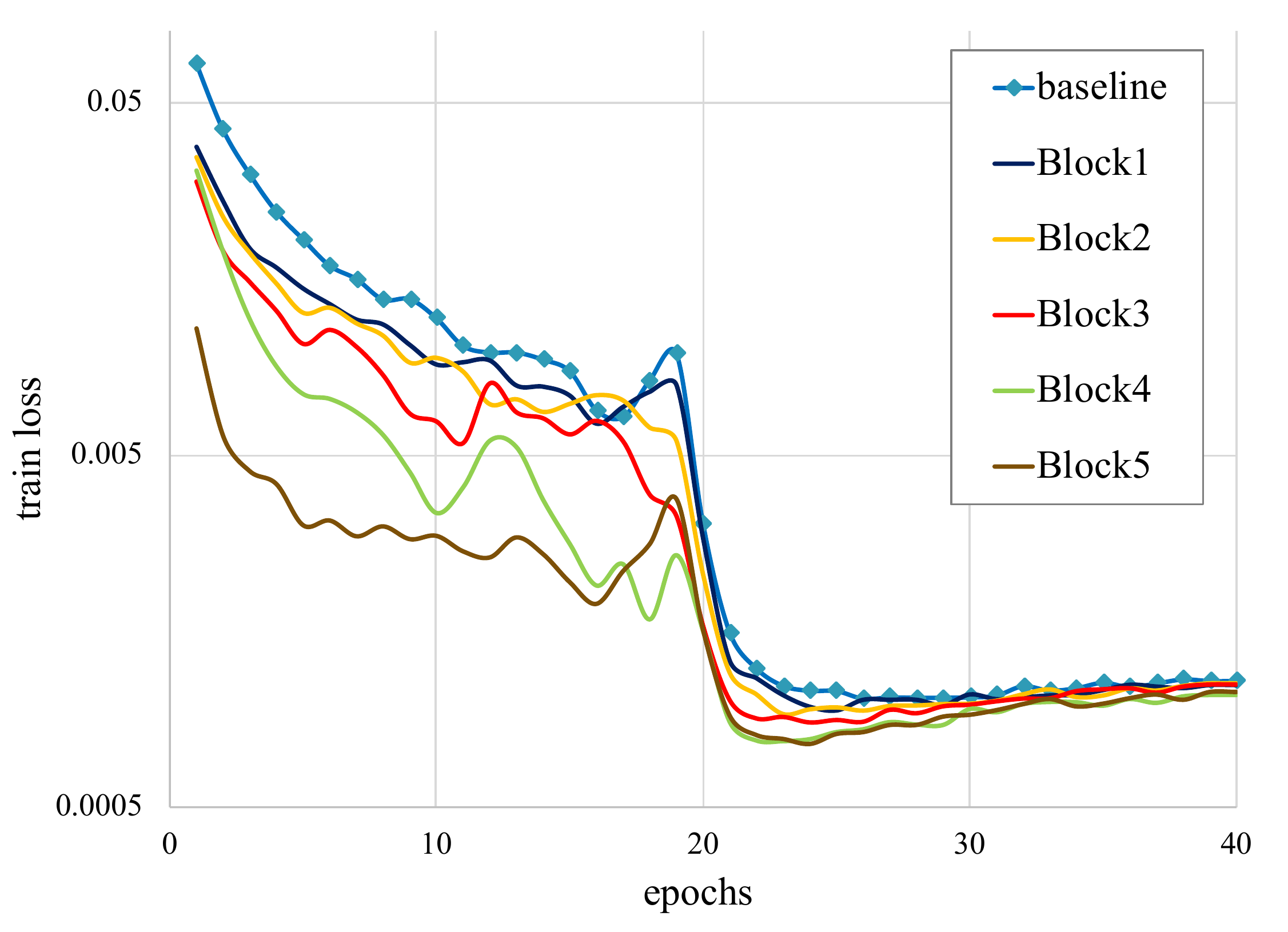}}
    \caption{The training loss convergences by ordinary fine-tuning (baseline) and rolling-back schemes where block $i$ is continuously tuned and the other blocks are rolled back to the pre-trained one.}
    \label{fig:oneblock}
\end{figure}
%%%%%%%%%%%%%%%%%%%%%%%%%%%%%%%%%%%%%%%%%%%%%%%%%%%%
%%%%%%%%%%%%%%%%%%%%%%%%%%%%%%%%%%%%%%%%%%%%%%%%%%%%
\begin{table}[t]
\centering
\caption{The generalization performance of each scheme in Figure  \ref{fig:oneblock}. Bold numbers show the best performance. }
\label{table:pretest}
\begin{tabular}{ccccc}
\toprule
remain layers & mAP  		& rank-1          & rank-5          & rank-10            \\  \midrule
baseline      & 73.16		& 89.43           & 96.35           & 97.77   \\
Block1        & 74.08	 & 89.49  & \textbf{96.50} & 97.62 \\
Block2        & \textbf{74.37} & \textbf{89.96} & \textbf{96.50} & 97.62\\
Block3        & 73.87		& 89.90 & 96.20          & \textbf{97.83}            \\
Block4        & 73.82		& 89.64          & 95.81          & 97.62    \\
Block5        & 71.17		& 88.45          & 95.61          & 97.42              \\ \bottomrule

\end{tabular}
\end{table}
%%%%%%%%%%%%%%%%%%%%%%%%%%%%%%%%%%%%%%%%%%%%%%%%%%%%
%%%%%%%%%%%%%%%%%%%%%%%%%%%%%%%%%%%%%%%%%%%%%%%%%%%%
\newcommand\tab[1][1.2cm]{\hspace*{#1}}
\renewcommand{\algorithmicrequire}{\textbf{Input:}}
\renewcommand{\algorithmicensure}{\textbf{Parameter:}}

\begin{algorithm*}
\caption{Re-fine learning}\label{alg:Re-fine}
\begin{algorithmic}[1]
\Ensure N: Number of total block , M: Number of lower block
\Ensure $\theta_{1}^{(0)},..,\theta_{N}^{(0)}$ : weights of pre-trained network
\Require $\theta_{1},..,\theta_{N}$ , $\theta_{FC}$, $X, Y$(dataset)

\State $\theta_i^{(1)} = \theta_i^{(0)},~ \forall~ i=1,.., N $ \Comment Initialize weights to pre-trained one
\State $\theta_{FC}^{(1)} \gets $ random initialization
\State $\hat{\theta}_{1}^{(1)},..,\hat{\theta}_{N}^{(1)},\hat{\theta}_{FC}^{(1)} \gets \textsc{Fine-Tune}(X,Y,\theta_{1}^{(1)},..,\theta_{N}^{(1)},\theta_{FC}^{(1)})$ \Comment First fine-tune on ReID dataset X,Y

\For{\texttt{p = 2 to M}} 
\State
  $\theta_i^{(p)} = 
   \begin{cases} 
      \hat{\theta_i}^{(p-1)} & i < p \\
      \theta_i^{(0)} & i\geq p 
   \end{cases}$ \Comment Remain certain layers and roll back others
\State 	$\theta_{FC}^{(p)} = \hat{\theta}_{FC}^{(p-1)}$ \Comment Do not roll back FC layers 
\State $\hat{\theta}_{1}^{(p)},..,\hat{\theta}_{N}^{(p)},\hat{\theta}_{FC}^{(p)} \gets \textsc{Fine-Tune}(X,Y,\theta_{1}^{(p)},..,\theta_{N}^{(p)},\theta_{FC}^{(p)})$ \Comment Refine-tune on ReID dataset X,Y
\EndFor
\end{algorithmic}
\end{algorithm*}
%%%%%%%%%%%%%%%%%%%%%%%%%%%%%%%%%%%%%%%%%%%%%%%%%%%%

\subsection{Analysis of fine-tuning method}
This section determines which layer converges insufficiently by conventional fine-tuning.
Figure~\ref{fig:oneblock} shows the convergence, supporting the key ideas of the proposed fine-tuning strategy. `baseline’ denotes the conventional fine-tuning, while `Block $i$' indicates the refine-tuning wherein every block except `Block $i$' is rolled back after the `baseline’ fine-tuning. 
Table~\ref{table:pretest} shows the generalization performance of each scheme.
{\bf A meaningful discovery} is that a rolling-back scheme with remaining low-level blocks (Block1, Block2, Block3) shows slower convergence than applying the rolling-back scheme to the remaining high-level blocks (Block3, Block4). 
However, as shown in Table~\ref{table:pretest}, the scheme that maintains the low-level blocks gives better generalization performance than the scheme preserving the high-level blocks. 
This indicates that the 'baseline' fine-tuning causes the low-level layers to be converged at a premature. 
This gives us an insight that rolling back of the high-level layers except the low-level layers might give the low-level layers an opportunity to learn further. 
As {\bf additional consideration}, all the weights cannot be given in pre-trained states. This is because the output layer of a deep network for a new task is usually different from the backbone network. 
Hence, the FC layers must be initialized in a random manner. 
Rolling back the FC layers to random states does not provide any benefit.
Thus, in our rolling-back scheme, FC layers are excluded from rolling back, although it is a high-level layer, to keep a consistent learning of the low-level layers. 

\subsection{Refine-tuning with rolling back}
The aforementioned analysis shows that a premature convergence degrades 
performance and rolling back high-level layers can be a beneficial strategy to mitigate the premature convergence problem in the low-level layers. 
For further tuning of the low-level layers, we designed a rolling-back refine-tuning scheme that trains the low-level layers incrementally from the front layer along with rolling back the remaining high-level layers.
The detailed rolling back scheme is described in the following.

\begin{enumerate}
\setcounter{enumi}{0}
\item In the first fine-tuning period~($p=1$), 
the weights, $\theta_{1},..,\theta_{N}$, are initialized with the pre-trained weights, ~$\theta_i^{(0)}$. 
\begin{equation}
	\theta_i^{(1)} = \theta_i^{(0)} , ~~\forall i=1, ..., N.
\end{equation}
The weights~($\theta_{FC}$) in FC layer are initialized with the random scratch~\cite{ref:he_initial}. Then the first period of fine-tuning is performed on the target dataset by Eq.~(\ref{eq:forwarding}), Eq.~(\ref{eq:ce-loss}). 
The updated weight of the $i$-th block is denoted by ~$\hat{\theta_i}^{(1)}$, which is obtained by minimizing the loss from Eq.~(\ref{eq:ce-loss}).

\item From the refine-tuning period with rolling back~($p\geq2$), we roll-back the high-level layers as in the following procedure. 
First, Block1~($\theta_{1}$) is maintained in the state of previous period and all the remaining blocks~($\theta_{2},...,\theta_{N}$) are rolled back to their pre-trained states $\theta_i^{(0)}$. In other words, Block1 continues the learning, and the other blocks restart the learning from the beginning with the pre-trained initial weights. In the incremental manner, the next low-level block is added one-by-one to the set of blocks continuing the learning, while the remaining ones are rolled back.
The rolling-back refine-tuning is repeated until all layers are included in the set of blocks continuing the learning. In summary, in the $p$-th refine-tuning period, the weights of the network are rolled back as

\begin{equation}
  \theta_i^{(p)} = 
   \begin{cases} 
      \hat{\theta_i}^{(p-1)} & i < p \\
      \theta_i^{(0)} & i \geq p, 
   \end{cases}
\end{equation}
where $\hat{\theta_i}^{(p-1)}, i=1, ..., N$ are the updated weights in the $(p-1)$-th refine-tuning period. 
During the refine-tuning process, the ($\theta_{FC}$) is not rolled back as mentioned above.
\begin{equation}
	\theta_{FC}^{(p)} = \hat{\theta}_{FC}^{(p-1)}. 
\end{equation}
\end{enumerate}
The detailed procedure of the refine-tuning scheme with rolling-back is summarized in Algorithm~\ref{alg:Re-fine}.

\section{Experiment}

\subsection{Dataset}
%%%%%%%%%%%%%%%%%%%%%%%%%%%%%%%%%%%%%%%%%%%%%%%%%%%%%%%%
\begin{table*}[t]
\centering
\caption{Results of our rolling-back scheme on different ReID dataset}
\vspace{-2mm}
\label{tabular:series}
\begin{tabular}{clcccccccc}
\toprule
\multicolumn{1}{l}{} & \multicolumn{1}{l}{} &\multicolumn{8}{c}{ResNet-50}\\
\multicolumn{1}{l}{} & \multicolumn{1}{l}{continuously} &\multicolumn{2}{c}{Market-1501} &\multicolumn{2}{c}{DukeMTMC}&\multicolumn{2}{c}{CUHK03-L}&\multicolumn{2}{c}{CUHK03-D}\\
 Period & tuned blocks    	& mAP 		& rank-1  & mAP  & rank-1  & mAP & rank-1& mAP & rank-1\\ \midrule
1	&none   				& 73.16 	& 89.43  & 63.26 & 80.83 & 45.17& 50.07 & 44.05& 48.00  \\
2	&B1+ FC      		    & 75.65 	& 90.95  & 66.09 & 81.96  & 47.69 & 51.21 & 45.76& 50.50 \\
3	&B1+B2+FC     			& 76.54    & 91.12   & \textbf{66.57}  & \textbf{82.41} & 49.98 & 54.36 & 46.20& 51.36 \\
4	&B1+B2+B3+FC &\textbf{77.01} &\textbf{91.24} & 66.39& 82.32  &\textbf{50.72}&\textbf{55.64} &\textbf{47.43}& \textbf{52.93}  \\ \bottomrule
\end{tabular}
\end{table*}
%%%%%%%%%%%%%%%%%%%%%%%%%%%%%%%%%%%%%%%%%%%%%%%%
\begin{table*}[t]
\centering
\caption{Results of our rolling-back scheme for different network types.}
\vspace{-2mm}
\label{tabular:family}
\begin{tabular}{clcccccccc}
\toprule
\multicolumn{1}{l}{} & \multicolumn{1}{l}{} &\multicolumn{4}{c}{ResNet-34}&\multicolumn{4}{c}{ResNet-101}\\
\multicolumn{1}{l}{} &  \multicolumn{1}{l}{continuously} &\multicolumn{2}{c}{Market-1501} &\multicolumn{2}{c}{DukeMTMC}&\multicolumn{2}{c}{Market-1501}&\multicolumn{2}{c}{DukeMTMC}\\
 Period & tuned blocks    	& mAP 		& rank-1  & mAP  & rank-1  & mAP & rank-1& mAP & rank-1\\ \midrule
1	&none   				& 70.65 	& 86.93  & 60.06 & 78.69 & 75.91& 90.80 & 66.00& 82.27  \\
2	&B1+ FC      		    & 73.63 	& 89.13  & 63.45 & 81.10  & 77.21 & 90.77 & 69.27& 83.62 \\
3	&B1+B2+FC     			& 74.85    & 90.02   & 65.16  & 82.18 & 78.17 & 91.27 & \textbf{70.24}& \textbf{85.19} \\
4	&B1+B2+B3+FC &\textbf{74.97} &\textbf{90.05} &\textbf{65.44} &\textbf{83.08} &\textbf{79.95} &\textbf{92.49}&69.88 &84.43  \\ \bottomrule
\end{tabular}
\label{ablation_test50}
\end{table*}
%%%%%%%%%%%%%%%%%%%%%%%%%%%%%%%%%%%%%%%%%%%%%%%%

\subsubsection{Market-1501} Market-1501~\cite{ref:market1501} is widely used dataset in person ReID. Market-1501 contains 32,668 images of 1,501 identities. All the bounding box images are results of detection by the DPM detector~\cite{ref:DPM}. The dataset is divided into a training set of 751 identities and a test set of 750 identities.
\subsubsection{DukeMTMC-ReID(DukeMTMC)} Based on the multi-target and multi-camera tracking dataset, DukeMTMC ~\cite{ref:SampleGAN} has been specially designed for person ReID. DukeMTMC contains 36,411 images of 1,402 identities which are   divided into a training set and a testing set of 702 and 702 identities, respectively.
\subsubsection{CUHK03-np} CUHK03-np~\cite{ref:re-rank2017} is a modified version of the original CUHK03 dataset. The hand-labeled (CUHK03-L) and DPM-detected~\cite{ref:DPM} bounding boxes (CUHK03-D) are offered. CUHK03-np contains 14,096 images of 1,467 identities. The new version is split into two balanced sets containing 767 and 700 identities for training and testing, respectively.

\vspace{-1mm}
\subsection{Implementation detail}
Our method was implemented using PyTorch~\cite{ref:pytorch} library. All inputs  are resized to 288$\times$144 and the batch size was set to 32. 
No other augmentation is used except horizontal flip in our training process. 
The initial learning rate was set to 0.01 and 0.1 for the feature extractor and the classifier, respectively. 
The learning rates were multiplied by 0.1 at every 20 epoch and we trained for 40 epochs as one refine-tuning period. 
In our experiment, the proposed refine-tuning strategy has been rolled back three times and four epochs have been trained for four refine-tuning periods, and so a total of 160 epochs are have been repeated for all fine-tuning. 
The learning rates of rolling back blocks are restored to 0.01 at the beginning of every period. In contrast, the blocks that do not roll back begin with the low learning rate of 0.001 since a high learning rate of the sufficiently trained blocks might yield sudden exploding.
The optimizer used in this study was stochastic gradient descent (SGD) with nesterov momentum~\cite{ref:nesterov}.
For the optimizer, the momentum rate and the weight decay were set to $0.9$ and $ 5 \times10 ^ {- 4} $, respectively. 
In every rolling back, the momentum of gradient was reset to $0$.
In the test process, the additional feature vector was used to add the feature vector of the horizontal flipped input pairwise. We report rank-1 accuracy of Cumulative Matching Characteristics (CMC) curve and the mean Average Precision (mAP) for performance evaluation.
%%%%%%%%%%%%%%%%%%%%%%%%%%%%%%%%%%%
\begin{figure*}[t]
\centering
{\includegraphics[width=18.0cm]{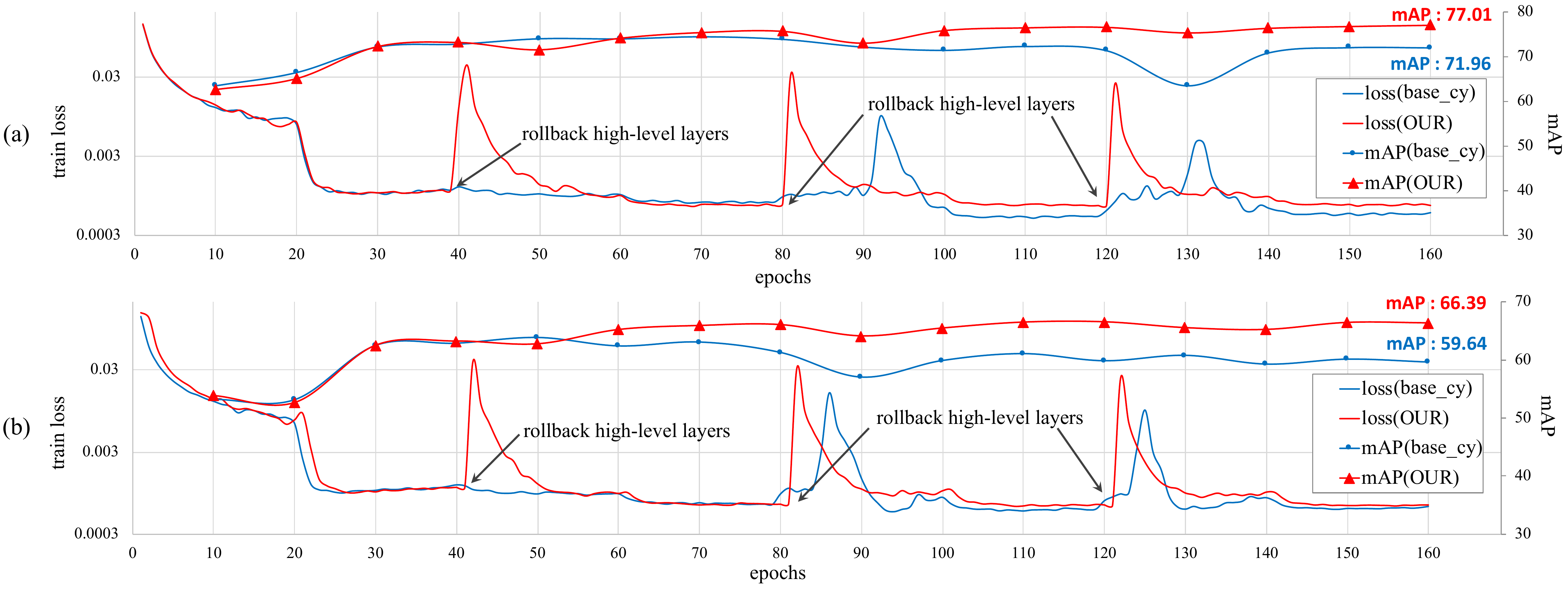}}
\caption{The train loss and mAP graph for comparison of our rolling-back scheme and the conventional fine-tuning at once. (a) is the results on Market-1501 and (b) is the results on DukeMTMC.}
    \label{fig:long1}
\end{figure*}
%%%%%%%%%%%%%%%%%%%%%%%%%%%%%%%

\vspace{-1mm}
\subsection{Ablation tests}
The network trained with the proposed strategy was verified via ablation tests on Market-1501, DukeMTMC, CUHK03-L and CUHK03-D.
The proposed refine-tuning strategy is applied to a network over four periods.
As the refine-tuning periods progress, the continuously tuned blocks are cumulative (e.g., B1+B2+FC in the third period). The other blocks are rolled back to their original pre-trained states.
As shown in Table~\ref{tabular:series}, the performance increases as the refine-tuning periods progress with the exception of DukeMTMC in the fourth period. 
However, even in this case, the gap was negligible. 
The improvement is most prominent in the second refine-tuning period during which the first rolling back is performed.  
To verify the generality of our refine-tuning scheme, we conducted additional experiments with other networks including ResNet-34 and ResNet-101~\cite{ref:resnet} under the same settings. 
Table~\ref{tabular:family} shows the performance of each network in Market-1501 and DukeMTMC. 
The proposed refine-tuning scheme also showed a consistent improvement in ResNet-34 and ResNet-101.
The ablation test results demonstrate that the proposed refine tuning scheme has a significant advantage as a general method to enhance the generalization performance in the ReID problem in which only a limited amount of data is available.

%%%%%%%%%%%%%%%%%%%%%%%
\begin{figure}[t]
\centering
{\includegraphics[width=8.5cm]{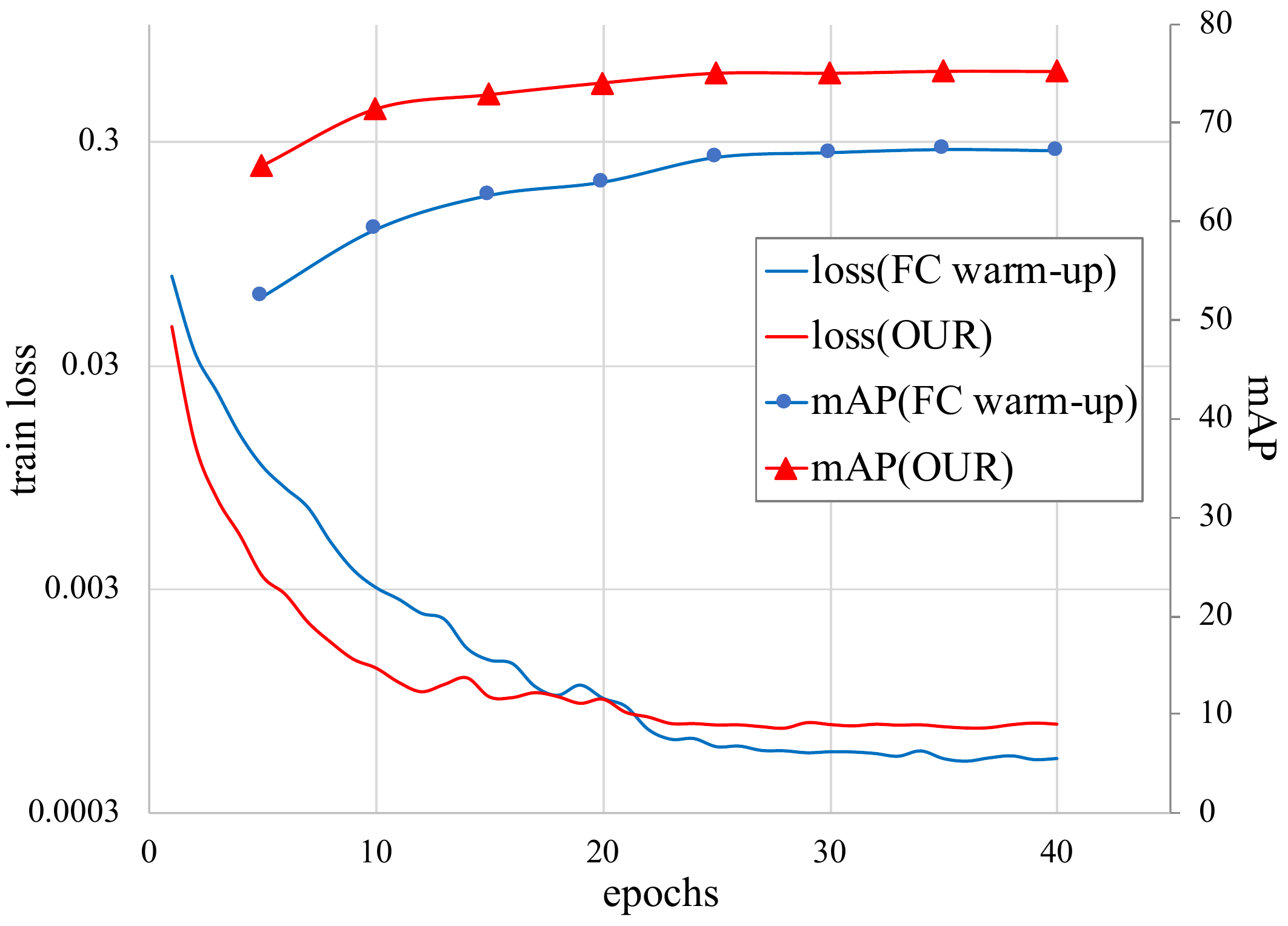}}
    \hfill    
    \caption{The results of comparison with FC warm-up training method}
    \label{fig:warmup}
\end{figure}
%%%%%%%%%%%%%%%%%%%%%%%%%%%%%
\begin{figure}[t]
\centering
{\includegraphics[width=8.0cm]{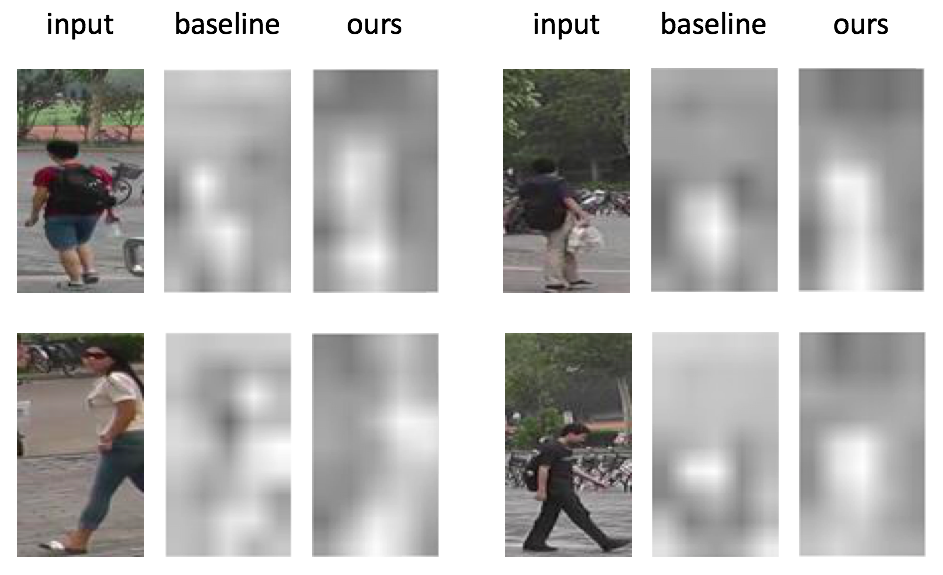}}
    \hfill    
    \caption{The attention maps formed by the last feature layer trained by our rolling-back scheme and the baseline}
    \label{fig:attention}
    \vspace{-0.4cm}
\end{figure}
%%%%%%%%%%%%%%%%%%%%%%%%%%%%%%%
\subsection{Effect of rolling back as a perturbation}
To evaluate the effect of our rolling-back scheme, it is compared with 'base$\_$cy' method that does roll back none of the block but merely adjusts the learning rate with the same timing as ours for a perturbation driving to other local basins.
The 'base$\_$cy' is similar to other studies~\cite{ref:warmSGD,ref:cyclical} that perturb only the learning rate.
Figure~\ref{fig:long1} shows the change in training loss and mAP of the whole processes of the proposed refine-tuning and the base$\_$cy fine-tuning.
After the first rolling-back at 40 epochs, the training loss from the rolling-back scheme converges to a value that is better than the value of the base$\_$cy in the 70-80 epochs. 
After the second and third rolling-backs, the training loss of the base$\_$cy converges to a lower value than that of the proposed method, but the base$\_$cy shows a worse generalization performance (mAP) than the proposed method. 

%%%%%%%%%%%%%%%%%%%%%%%%%%%%%%%%%%%%%%%%%%%%%%%%
\begin{table*}[t]
\centering
\caption{Comparison with State-of-the-art methods on Market-1501, DukeMTMC and CUHK03-L/D}
\label{tabular:sota_compare}
\begin{tabular}{rlcccccccccc}
\toprule
\multicolumn{1}{l}{} & \multicolumn{1}{l}{} &\multicolumn{2}{c}{Market-1501}& \multicolumn{2}{c}{DukeMTMC}  & \multicolumn{2}{c}{CUHK03-L}  & \multicolumn{2}{c}{CUHK03-D}        \\ 
Method  							& Backbone  & mAP  & rank-1 &	 mAP  & rank-1 & mAP  & rank-1 &	 mAP  & rank-1&  Add-on     \\ \midrule
PT-GAN~\shortcite{ref:PoseTransfer_GAN}& ResNet-50  & 58.0 & 79.8 & 	48.1 &  68.6  & 30.5& 33.8& 28.2& 30.1& pose+GAN        \\

SVDNet~\shortcite{ref:SVDnet}     & ResNet-50  & 62.1 & 82.3   &     56.8 &  76.7  & 37.8&40.9 &37.3 & 41.5&   -         \\
PDC~\shortcite{ref:pose_driven}   & Inception&  63.4 & 84.1   &   	 -   &    -  &- &- &- &- &  pose       \\
AACN~\shortcite{ref:AACN}    		& GoogleNet & 66.9 & 85.9  &  	59.3  & 76.8   & -&- &- &- & pose   \\
HAP2S\_P~\shortcite{ref:Hard-aware}    		  & ResNet-50  & 69.4 & 84.6 &   	 60.6   &    75.9  & -& -& -&- &  -         \\
PSE~\shortcite{ref:Pose_sensitive}  & ResNet    & 69.0 & 87.7   &	62.0  &  79.8   & -& -& -& - & pose       \\
CamStyle~\shortcite{ref:CamAug} & ResNet-50  & 71.6 & 89.2  & 		57.6 &  78.3   & -&- &- &- & GAN\\
PN-GAN~\shortcite{ref:PoseNormal_GAN}& ResNet-50  & 72.6 & 89.4 & 	53.2 &  73.6   & -& -& -& -& pose+GAN        \\
MGCAM~\shortcite{ref:MaskreID}    & MSCAN     & 74.3 & 83.8  &   	 -    &    -   & 50.2& 50.1& 46.7&46.9 &   mask       \\ 
MLFN~\shortcite{ref:MLFN}     & Original& 74.3 & 90.0  & 			62.8 &  81.0    & 49.2&54.7 &47.8 &52.8 & -         \\
HA-CNN~\shortcite{ref:harmonious}   & Inception & 75.7 & 91.2  & 	63.8 &  80.5    &41.0 &44.4 &38.6 &41.7 & -          \\
DuATM~\shortcite{ref:DuATM}    & DenseNet-121  & 76.6 & 91.4 &  64.6  &81.8   &-& -&- &- &    -         \\ \midrule
Ours    						  & ResNet-34  &  75.0 & 90.1  &   65.4 &  83.1   & 48.6&53.0 &45.6 & 51.3&   -       \\
Ours   							  & ResNet-50  &  77.0 & 91.2  & 	66.6 &  82.4   & 50.7&55.6 &47.4 &52.9 &   -       \\
Ours   				  & ResNet-101&\textbf{79.9}&\textbf{92.5}  & 	\textbf{70.2} &  \textbf{85.2}  & \textbf{55.7} &\textbf{59.8}& \textbf{50.5}&  \textbf{55.6}&   -           \\ \bottomrule 
\end{tabular}
\end{table*}
%%%%%%%%%%%%%%%%%%%%%%%%%%%%%%%%%%%%%%%%%%%%%%%%
%%%%%%%%%%%%%%%%%%%%%%%%%%%%%%%%%%%%%%%%%%%%%%%%
\newcolumntype{L}[1]{>{\raggedright\let\newline\\\arraybackslash\hspace{0pt}}m{#1}}
\newcolumntype{C}[1]{>{\centering\let\newline\\\arraybackslash\hspace{0pt}}m{#1}}
\newcolumntype{R}[1]{>{\raggedleft\let\newline\\\arraybackslash\hspace{0pt}}m{#1}}

\begin{table}[t]
\centering
\caption{Comparison with State-of-the-art methods using same backbone network ResNet-50}
\label{tabular:sota_res50}
\begin{tabular}{rlcccc}
%\begin{tabular}{|R|L|C{2cm}|C|C|C|C|C|}
\toprule
\multicolumn{1}{l}{}  &\multicolumn{2}{c}{Market-1501}& \multicolumn{2}{c}{DukeMTMC}        \\ 
Method  							&  mAP  & rank-1 &	 mAP  & rank-1  & Add-on     \\ \midrule
PT-GAN&  58.0 & 79.8 & 	48.1 &  68.6      &GAN  \\
SVDNet     &  62.1 & 82.3   &     56.8 &  76.7   &   -        \\
HAP2S\_P  &  69.4 & 84.6  &   	 60.6   &   75.9  &   -         \\
CamStyle &  71.6 & 89.2  & 		57.6 &  78.3 &GAN \\
PN-GAN&  72.6 & 89.4 & 	53.2 &  73.6         &GAN \\ \midrule
Ours   			&   \textbf{77.0} & \textbf{91.2}  & 	\textbf{66.6} &  \textbf{82.4}  & -   \\ \bottomrule 
\end{tabular}
\end{table}
%%%%%%%%%%%%%%%%%%%%%%%%%%%%%%%%%%%%%%%%%%%%%%%%

\subsection{Comparison to FC warm-up training}
In this section, we discuss the difference between our method and FC warm-up training~\cite{ref:resnet}.
As mentioned previously, the new FC layers start randomly from scratch. 
FC warm-up is a way to freeze the pre-trained weights in all hidden layers except for the FC layers and train the FC layers before starting the main fine-tuning. 
In the comparison experiment, the baseline was warmed up for 20 epochs. 
In our proposed method, period 1 (see Table~\ref{tabular:series}) is similar to FC warm-up where FC layers start from random scratch.
However, the proposed method does not freeze the pre-trained weights in period 1. 
The training loss and mAP for FC warm-up and our methods
are depicted in Figure~\ref{fig:warmup}. 
FC warm-up and our methods start fine/refine-tuning after training the FC layers.
The FC warm-up converges to a lower training loss than the proposed method, but the proposed method shows better performance in terms of generalization.

\subsection{Attention performance of our refine-tuning method}
To learn discriminative features for the ReID task, it is important to distinguish the foreground from the background.
Figure~\ref{fig:attention} shows that our method can generate a more distinguishable feature map in the last convolutional layer than the baseline of the conventional fine-tuning method.

\subsection{Comparisons with state-of-the-art methods}
We also compared the proposed method with state-of-the-art methods. 
Table~\ref{tabular:sota_res50} shows the comparison results when using ResNet-50. 
The proposed rolling-back refine-tuning scheme shows the best performance even though our method does not use any add-on scheme. 
Furthermore, compared to other methods without add-on scheme (SVDNet, HAP2S\_P), our method outperforms them by more than 7\% mAP improvement for Market-1501. 
Table~\ref{tabular:sota_compare} summarizes the results compared with the state-of-the-art methods on Market-1501, DukeMCMT, and CUHK03-L/D. According to the results, the rolling-back refine-tuning scheme makes a meaningful contribution to the enhancement of any backbone networks so that it outperforms state-of-the-art algorithms utilizing add-on schemes.

\section{Conclusion}
In this paper, we proposed a refine tuning method with a rolling-back scheme which further enhances the backbone network. The key idea of the rolling-back scheme is to restore the weights in a part of the backbone network to the pre-trained weights when the fine-tuning converges at a premature state.
To escape from the premature state, we adopt an incremental refine tuning strategy by applying the fine tuning repeatedly, along with the rolling-back. According to the experimental results, the rolling-back scheme makes a meaningful contribution to enhancement of the backbone network where it derives the convergence to a local basin of a good generalization performance. As a result, our method without any add-on scheme could outperform the state-of-the-arts with help of add-on scheme.

\section{Acknowledgement}
This work was supported by  
Next-Generation ICD Program through NRF funded by Ministry of S\&ICT [2017M3C4A7077582], 
ICT R\&D program MSIP/IITP [2017-0-00306, Outdoor Surveillance Robots].

\bibliographystyle{aaai}
\bibliography{refs}

\begin{thebibliography}{}

\bibitem[\protect\citeauthoryear{Bossard, Guillaumin, and
  Van~Gool}{}]{ref:food-101}
Bossard, L.; Guillaumin, M.; and Van~Gool, L.
\newblock Food-101--mining discriminative components with random forests.
\newblock In {\em ECCV, pages={446--461}, year={2014},
  organization={Springer}}.

\bibitem[\protect\citeauthoryear{Cao \bgroup et al\mbox.\egroup
  }{2017}]{ref:openpose}
Cao, Z.; Simon, T.; Wei, S.-E.; and Sheikh, Y.
\newblock 2017.
\newblock Realtime multi-person 2d pose estimation using part affinity fields.
\newblock In {\em CVPR}.

\bibitem[\protect\citeauthoryear{Chang, Hospedales, and Xiang}{2018}]{ref:MLFN}
Chang, X.; Hospedales, T.~M.; and Xiang, T.
\newblock 2018.
\newblock Multi-level factorisation net for person re-identification.
\newblock In {\em CVPR}.

\bibitem[\protect\citeauthoryear{Chen \bgroup et al\mbox.\egroup
  }{2017}]{ref:beyond_triplet}
Chen, W.; Chen, X.; Zhang, J.; and Huang, K.
\newblock 2017.
\newblock Beyond triplet loss: a deep quadruplet network for person
  re-identification.
\newblock In {\em CVPR}.

\bibitem[\protect\citeauthoryear{Deng \bgroup et al\mbox.\egroup
  }{2009}]{ref:imagenet}
Deng, J.; Dong, W.; Socher, R.; Li, L.-J.; Li, K.; and Fei-Fei, L.
\newblock 2009.
\newblock Imagenet: A large-scale hierarchical image database.
\newblock In {\em CVPR}.
\newblock Ieee.

\bibitem[\protect\citeauthoryear{Felzenszwalb \bgroup et al\mbox.\egroup
  }{2010}]{ref:DPM}
Felzenszwalb, P.~F.; Girshick, R.~B.; McAllester, D.; and Ramanan, D.
\newblock 2010.
\newblock Object detection with discriminatively trained part-based models.
\newblock {\em IEEE Trans. on PAMI}.

\bibitem[\protect\citeauthoryear{He \bgroup et al\mbox.\egroup
  }{2015}]{ref:he_initial}
He, K.; Zhang, X.; Ren, S.; and Sun, J.
\newblock 2015.
\newblock Delving deep into rectifiers: Surpassing human-level performance on
  imagenet classification.
\newblock In {\em ICCV}.

\bibitem[\protect\citeauthoryear{He \bgroup et al\mbox.\egroup
  }{2016}]{ref:resnet}
He, K.; Zhang, X.; Ren, S.; and Sun, J.
\newblock 2016.
\newblock Deep residual learning for image recognition.
\newblock In {\em CVPR}.

\bibitem[\protect\citeauthoryear{Hermans, Beyer, and
  Leibe}{2017}]{ref:defense_triplet}
Hermans, A.; Beyer, L.; and Leibe, B.
\newblock 2017.
\newblock In defense of the triplet loss for person re-identification.
\newblock {\em arXiv preprint arXiv:1703.07737}.

\bibitem[\protect\citeauthoryear{Hinton, Vinyals, and
  Dean}{2015}]{ref:distillation}
Hinton, G.; Vinyals, O.; and Dean, J.
\newblock 2015.
\newblock Distilling the knowledge in a neural network.
\newblock {\em arXiv preprint arXiv:1503.02531}.

\bibitem[\protect\citeauthoryear{Insafutdinov \bgroup et al\mbox.\egroup
  }{2016}]{ref:deepcut_pose}
Insafutdinov, E.; Pishchulin, L.; Andres, B.; Andriluka, M.; and Schiele, B.
\newblock 2016.
\newblock Deepercut: A deeper, stronger, and faster multi-person pose
  estimation model.
\newblock In {\em ECCV}.

\bibitem[\protect\citeauthoryear{Kalayeh \bgroup et al\mbox.\egroup
  }{2018}]{ref:human_parsing}
Kalayeh, M.~M.; Basaran, E.; Gokmen, M.; Kamasak, M.~E.; and Shah, M.
\newblock 2018.
\newblock Human semantic parsing for person re-identification.
\newblock In {\em CVPR}.

\bibitem[\protect\citeauthoryear{Koestinger \bgroup et al\mbox.\egroup
  }{2012}]{ref:metric_learning_reID}
Koestinger, M.; Hirzer, M.; Wohlhart, P.; Roth, P.~M.; and Bischof, H.
\newblock 2012.
\newblock Large scale metric learning from equivalence constraints.
\newblock In {\em CVPR}.

\bibitem[\protect\citeauthoryear{Kornblith, Shlens, and
  Le}{2018}]{ref:do_better_imagenet}
Kornblith, S.; Shlens, J.; and Le, Q.~V.
\newblock 2018.
\newblock Do better imagenet models transfer better?
\newblock {\em arXiv preprint arXiv:1805.08974}.

\bibitem[\protect\citeauthoryear{Kuo, Khamis, and
  Shet}{2013}]{ref:color_hist_wacv}
Kuo, C.-H.; Khamis, S.; and Shet, V.
\newblock 2013.
\newblock Person re-identification using semantic color names and rankboost.
\newblock In {\em WACV}.

\bibitem[\protect\citeauthoryear{Li and
  Hoiem}{2017}]{ref:learning_without_forgetting}
Li, Z., and Hoiem, D.
\newblock 2017.
\newblock Learning without forgetting.
\newblock {\em IEEE Transactions on Pattern Analysis and Machine Intelligence}.

\bibitem[\protect\citeauthoryear{Li, Zhu, and Gong}{2018}]{ref:harmonious}
Li, W.; Zhu, X.; and Gong, S.
\newblock 2018.
\newblock Harmonious attention network for person re-identification.
\newblock In {\em CVPR}.

\bibitem[\protect\citeauthoryear{Liu \bgroup et al\mbox.\egroup
  }{2018}]{ref:PoseTransfer_GAN}
Liu, J.; Ni, B.; Yan, Y.; Zhou, P.; Cheng, S.; and Hu, J.
\newblock 2018.
\newblock Pose transferrable person re-identification.
\newblock In {\em CVPR}.

\bibitem[\protect\citeauthoryear{Long, Shelhamer, and Darrell}{2015}]{ref:FCN}
Long, J.; Shelhamer, E.; and Darrell, T.
\newblock 2015.
\newblock Fully convolutional networks for semantic segmentation.
\newblock In {\em CVPR}.

\bibitem[\protect\citeauthoryear{Loshchilov and Hutter}{2016}]{ref:warmSGD}
Loshchilov, I., and Hutter, F.
\newblock 2016.
\newblock Sgdr: Stochastic gradient descent with warm restarts.
\newblock {\em arXiv preprint arXiv:1608.03983}.

\bibitem[\protect\citeauthoryear{Mahendran and
  Vedaldi}{2015}]{ref:understand_deepimage}
Mahendran, A., and Vedaldi, A.
\newblock 2015.
\newblock Understanding deep image representations by inverting them.
\newblock In {\em CVPR}.

\bibitem[\protect\citeauthoryear{Maji \bgroup et al\mbox.\egroup
  }{2013}]{ref:Aircraft}
Maji, S.; Rahtu, E.; Kannala, J.; Blaschko, M.; and Vedaldi, A.
\newblock 2013.
\newblock Fine-grained visual classification of aircraft.
\newblock {\em arXiv preprint arXiv:1306.5151}.

\bibitem[\protect\citeauthoryear{Nesterov}{1983}]{ref:nesterov}
Nesterov, Y.
\newblock 1983.
\newblock A method for unconstrained convex minimization problem with the rate
  of convergence o (1/k\^{} 2).
\newblock In {\em Doklady AN USSR}.

\bibitem[\protect\citeauthoryear{Paszke \bgroup et al\mbox.\egroup
  }{2017}]{ref:pytorch}
Paszke, A.; Gross, S.; Chintala, S.; Chanan, G.; Yang, E.; DeVito, Z.; Lin, Z.;
  Desmaison, A.; Antiga, L.; and Lerer, A.
\newblock 2017.
\newblock Automatic differentiation in pytorch.

\bibitem[\protect\citeauthoryear{Pedagadi \bgroup et al\mbox.\egroup
  }{2013}]{ref:color_hist_ReID}
Pedagadi, S.; Orwell, J.; Velastin, S.; and Boghossian, B.
\newblock 2013.
\newblock Local fisher discriminant analysis for pedestrian re-identification.
\newblock In {\em CVPR}.

\bibitem[\protect\citeauthoryear{Qian \bgroup et al\mbox.\egroup
  }{2018}]{ref:PoseNormal_GAN}
Qian, X.; Fu, Y.; Wang, W.; Xiang, T.; Wu, Y.; Jiang, Y.-G.; and Xue, X.
\newblock 2018.
\newblock Pose-normalized image generation for person re-identification.
\newblock In {\em ECCV}.

\bibitem[\protect\citeauthoryear{Ren \bgroup et al\mbox.\egroup
  }{2015}]{ref:faster}
Ren, S.; He, K.; Girshick, R.; and Sun, J.
\newblock 2015.
\newblock Faster r-cnn: Towards real-time object detection with region proposal
  networks.
\newblock In {\em NIPS}.

\bibitem[\protect\citeauthoryear{Sarfraz \bgroup et al\mbox.\egroup
  }{2018}]{ref:Pose_sensitive}
Sarfraz, M.~S.; Schumann, A.; Eberle, A.; and Stiefelhagen, R.
\newblock 2018.
\newblock A pose-sensitive embedding for person re-identification with expanded
  cross neighborhood re-ranking.
\newblock In {\em CVPR}.

\bibitem[\protect\citeauthoryear{Si \bgroup et al\mbox.\egroup
  }{2018}]{ref:DuATM}
Si, J.; Zhang, H.; Li, C.-G.; Kuen, J.; Kong, X.; Kot, A.~C.; and Wang, G.
\newblock 2018.
\newblock Dual attention matching network for context-aware feature sequence
  based person re-identification.
\newblock {\em arXiv preprint arXiv:1803.09937}.

\bibitem[\protect\citeauthoryear{Smith}{2017}]{ref:cyclical}
Smith, L.~N.
\newblock 2017.
\newblock Cyclical learning rates for training neural networks.
\newblock In {\em WACV}.

\bibitem[\protect\citeauthoryear{Song \bgroup et al\mbox.\egroup
  }{2018}]{ref:MaskreID}
Song, C.; Huang, Y.; Ouyang, W.; and Wang, L.
\newblock 2018.
\newblock Mask-guided contrastive attention model for person re-identification.
\newblock In {\em CVPR}.

\bibitem[\protect\citeauthoryear{Su \bgroup et al\mbox.\egroup
  }{2017}]{ref:pose_driven}
Su, C.; Li, J.; Zhang, S.; Xing, J.; Gao, W.; and Tian, Q.
\newblock 2017.
\newblock Pose-driven deep convolutional model for person re-identification.
\newblock In {\em ICCV}.

\bibitem[\protect\citeauthoryear{Sun \bgroup et al\mbox.\egroup
  }{2017}]{ref:SVDnet}
Sun, Y.; Zheng, L.; Deng, W.; and Wang, S.
\newblock 2017.
\newblock Svdnet for pedestrian retrieval.
\newblock In {\em ICCV}.

\bibitem[\protect\citeauthoryear{Szegedy \bgroup et al\mbox.\egroup
  }{2016}]{ref:InceptionV3}
Szegedy, C.; Vanhoucke, V.; Ioffe, S.; Shlens, J.; and Wojna, Z.
\newblock 2016.
\newblock Rethinking the inception architecture for computer vision.
\newblock In {\em Proceedings of the IEEE conference on computer vision and
  pattern recognition},  2818--2826.

\bibitem[\protect\citeauthoryear{Wah \bgroup et al\mbox.\egroup
  }{2011}]{ref:CUB-2011}
Wah, C.; Branson, S.; Welinder, P.; Perona, P.; and Belongie, S.
\newblock 2011.
\newblock The caltech-ucsd birds-200-2011 dataset.

\bibitem[\protect\citeauthoryear{Xu \bgroup et al\mbox.\egroup
  }{2018}]{ref:AACN}
Xu, J.; Zhao, R.; Zhu, F.; Wang, H.; and Ouyang, W.
\newblock 2018.
\newblock Attention-aware compositional network for person re-identification.
\newblock {\em arXiv preprint arXiv:1805.03344}.

\bibitem[\protect\citeauthoryear{Yu \bgroup et al\mbox.\egroup
  }{2018}]{ref:Hard-aware}
Yu, R.; Dou, Z.; Bai, S.; Zhang, Z.; Xu, Y.; and Bai, X.
\newblock 2018.
\newblock Hard-aware point-to-set deep metric for person re-identification.
\newblock In {\em ECCV}.

\bibitem[\protect\citeauthoryear{Zeiler and Fergus}{2014}]{ref:zfnet}
Zeiler, M.~D., and Fergus, R.
\newblock 2014.
\newblock Visualizing and understanding convolutional networks.
\newblock In {\em European conference on computer vision}.

\bibitem[\protect\citeauthoryear{Zhang \bgroup et al\mbox.\egroup
  }{2017}]{ref:mutualReID}
Zhang, Y.; Xiang, T.; Hospedales, T.~M.; and Lu, H.
\newblock 2017.
\newblock Deep mutual learning.
\newblock {\em arXiv preprint arXiv:1706.00384}.

\bibitem[\protect\citeauthoryear{Zhao \bgroup et al\mbox.\egroup
  }{2017}]{ref:pose_alined}
Zhao, L.; Li, X.; Zhuang, Y.; and Wang, J.
\newblock 2017.
\newblock Deeply-learned part-aligned representations for person
  re-identification.
\newblock In {\em ICCV}.

\bibitem[\protect\citeauthoryear{Zheng \bgroup et al\mbox.\egroup
  }{2015}]{ref:market1501}
Zheng, L.; Shen, L.; Tian, L.; Wang, S.; Wang, J.; and Tian, Q.
\newblock 2015.
\newblock Scalable person re-identification: A benchmark.
\newblock In {\em Computer Vision, IEEE International Conference on}.

\bibitem[\protect\citeauthoryear{Zheng, Yang, and
  Hauptmann}{2016}]{ref:ReID_review}
Zheng, L.; Yang, Y.; and Hauptmann, A.~G.
\newblock 2016.
\newblock Person re-identification: Past, present and future.
\newblock {\em arXiv preprint arXiv:1610.02984}.

\bibitem[\protect\citeauthoryear{Zheng, Zheng, and Yang}{2017}]{ref:SampleGAN}
Zheng, Z.; Zheng, L.; and Yang, Y.
\newblock 2017.
\newblock Unlabeled samples generated by gan improve the person
  re-identification baseline in vitro.
\newblock In {\em ICCV}.

\bibitem[\protect\citeauthoryear{Zhong \bgroup et al\mbox.\egroup
  }{2017a}]{ref:re-rank2017}
Zhong, Z.; Zheng, L.; Cao, D.; and Li, S.
\newblock 2017a.
\newblock Re-ranking person re-identification with k-reciprocal encoding.
\newblock In {\em CVPR}.

\bibitem[\protect\citeauthoryear{Zhong \bgroup et al\mbox.\egroup
  }{2017b}]{ref:pert_erasing}
Zhong, Z.; Zheng, L.; Kang, G.; Li, S.; and Yang, Y.
\newblock 2017b.
\newblock Random erasing data augmentation.
\newblock {\em arXiv preprint arXiv:1708.04896}.

\bibitem[\protect\citeauthoryear{Zhong \bgroup et al\mbox.\egroup
  }{2018}]{ref:CamAug}
Zhong, Z.; Zheng, L.; Zheng, Z.; Li, S.; and Yang, Y.
\newblock 2018.
\newblock Camera style adaptation for person re-identification.
\newblock In {\em CVPR}.

\end{thebibliography}

\end{document}